# A Cyber-Physical Architecture for Microgrids based on Deep learning and LORA Technology

Mojtaba Mohammadi, Abdollah Kavousi-Fard, Senior member, *IEEE*, Mortza Dabbaghjamanesh, Senior member, *IEEE, Mostafa Shaaban, Senior member, IEEE, Hatem. H. Zeineldin, Senior member, IEEE, Ehab Fahmy El-Saadany, Fellow member, IEEE*

*Abstract*— This paper proposes a cyber-physical architecture for the secured social operation of isolated hybrid microgrids (HMGs). On the physical side of the proposed architecture, an optimal scheduling scheme considering various renewable energy sources (RESs) and fossil fuel-based distributed generation units (DGs) is proposed. Regarding the cyber layer of MGs, a wireless architecture based on low range wide area (LORA) technology is introduced for advanced metering infrastructure (AMI) in smart electricity grids. In the proposed architecture, the LORA data frame is described in detail and designed for the application of smart meters considering DGs and ac-dc converters. Additionally, since the cyber layer of smart grids is highly vulnerable to cyber-attacks, t¹his paper proposes a deep-learning-based cyber-attack detection model (CADM) based on bidirectional long short-term memory (BLSTM) and sequential hypothesis testing (SHT) to detect false data injection attacks (FDIA) on the smart meters within AMI. The performance of the proposed energy management architecture is evaluated using the IEEE 33-bus test system. In order to investigate the effect of FDIA on the isolated HMGs and highlight the interactions between the cyber layer and physical layer, an FDIA is launched against the test system. The results showed that a successful attack can highly damage the system and cause widespread load shedding. Also, the performance of the proposed CADM is examined using a real-world dataset. Results prove the effectiveness of the proposed CADM in detecting the attacks using only two samples.

NOMENCLATURE

| | |
|---|---|
| $C_{Gi}^t$ | Generation cost of DGs |
| $b_i, b_f, b_o, \ldots$ | Bias vectors of the LSTM cell |
| $c_t$ | Cell state at time $t$ |
| $DR_i/UR_i$ | DGs' Ramp down/up rate |
| $f(S\|H)$ | Conditional mass function |
| $f_t$ | LSTM's Forget gate at time $t$ |
| $h(X)$ | Cost objective function |
| $i_t$ | Input gate at time $t$ |
| $m$ | Number of ones in the sample set |
| $n$ | Sample set size |
| $N_i$ | Number of time intervals |
| $N_d/N_{d\text{-}dc}/N_{d\text{-}ac}$ | Number of DGs in the MG/dc sub-grid/ac sub-grid |
| $N_{Load\text{-}dc}$ | Number of loads in dc sub-grid |
| $N_b$ | Number of buses |
| $N_{Load}$ | Number of loads in the MG |
| $o_t$ | Output gate at time $t$ |
| $P_{load\text{-}dc}$ | The active power demand in the dc sub-grid |
| $P^{inj,t}_j / Q^{inj,t}_j$ | Active/reactive power injected to the $j^{th}$ bus |
| $P_{Gi}^{min} / P_{Gi}^{max}$ | Min/max DGs' active power |
| $P_{conv}^{min} / P_{conv}^{max}$ | Min/max power of the ac-dc converter |
| $P_{i,max}^{line,t}$ | Feeder's capacity |
| $P_{lm}^i / P_{lf}^i$ | Measured/forecasted load value of the $i^{th}$ meter |
| $P_{Loss\text{-}dc}$ | Power loss in dc-subgrid |
| $RES^t$ | The spinning reserve at interval $t$ |
| $S_{Gi}^{on}/S_{Gi}^{off}$ | DGs' startup/shutdown cost |
| $S_l^i$ | Binary sample of the $i^{th}$ smart meter related to the CADM |
| $u_i^t$ | ON/OFF status of DGs |
| $UE^i/LE^i$ | Upper/lower forecasting error bounds |
| $U^i/L^i$ | Upper/lower bound of the SHT |
| $V_{min}^i/V_{max}^i$ | Min/max voltage of the $i^{th}$ bus |
| $V/\delta$ | Magnitude/phase of the voltage |
| $W_{f,x}, W_{i,x}, W_{o,h}, \ldots$ | Weighting matrices of the LSTM cell |
| $X$ | Control variables |
| $Y/\Theta$ | Magnitude/phase of the bus admittance matrix elements |
| $\alpha_i$ | User-selected false positive value |
| $\beta_i$ | User-selected false negative value |
| $\sigma$ | Logistic sigmoid function |

## I. INTRODUCTION

Generally, microgrids (MGs) are self-governing systems that may include RESs, loads, DGs, etc. that can operate either in the grid-connected or isolated mode. From the voltage type standpoint, MGs are categorized into three categories: ac, dc, and hybrid type. The concept of hybrid MG is proposed to achieve benefits associated with both ac and dc MGs at the same time. MGs are complicated cyber-physical systems that

M. Mohammadi, A. Kavousifard, and M Dabbaghjamanesh are with the Department of Electrical and Electronics Engineering, Shiraz University of Technology, Shiraz, Iran (Email: mojtabamohammadi303@gmail.com, kavousi@sutech.ac.ir, dabaghmanesh.morteza@gmail.com).

mainly include two layers. The first layer is the physical layer that is responsible for providing proper electrical power flow in the system, including physical devices such as transmission lines, power generation units, protection devices, energy storage devices, substations, and loads. The second layer is the cyber layer, which includes advanced metering infrastructure (AMI), communication platforms, and data management systems that are responsible for providing data flow in the system and communicating between the different parts of the grid [1-3]. Since the reliable and efficient operation of the MG is tied with its secured and optimal operation, security and management of both layers must be taken into account, carefully. In terms of operation and energy management of the system, a novel energy management framework has been proposed in this paper. The proposed optimization method is essential for efficiently managing and operating the microgrid system. By considering factors such as energy demand, generation capacity, storage capabilities, and grid connection, the method optimally schedules generation and storage resources, coordinates energy exchange with the main grid, and enhances load balancing and renewable energy integration. It improves system performance, cost-effectiveness, reliability, and environmental sustainability, serving as a valuable tool for microgrid operators to make informed decisions and optimize resource utilization.

Significant research effort has been conducted on the physical management of the MGs in recent years. In [4], a stochastic energy management scheme for grid-connected hybrid MGs (HMGs) considering the RESs, as well as uncertainties in the system is proposed. A secured deterministic management scheme for the optimal management of isolated as well as grid-connected hybrid MGs is introduced in [5]. Considering the physical and cyber operation of the system individually is the main drawback of [5]. The approach, in [6], proposed a multi-agent management scheme based on the cloud-fog approach and consensus optimization algorithm for the optimal operation of MGs. A two-level hierarchical energy management framework was proposed in [7] for the operation of MGs in two levels, wherein the lower level is responsible to manage the MG and the upper-level deals with the upstream grid and MG community level devices. A multi-level management scheme for HMGs considering RESs, ac, and dc loads, and dispatchable DGs was presented in [8]. In [9] a decentralized power flow control and management framework for the optimal operation of HMGs is introduced. It is worth noting that the above works [6-9] have addressed only the physical and electrical operation of the MG and have not considered the grid as a cyber-physical system. This issue is addressed in this study.

In smart electricity grids, the control and optimization of the system are carried out using data collected using metering devices. Therefore, the optimal and intelligent operation of MGs is tied with secured and reliable monitoring of the system. The AMI is the key technology that provides bidirectional communication between the MG's central control and end-users. While some researches on the physical layer of MGs is mentioned above, several pieces of research focusing on the architecture of the AMI considering different methods and technologies are investigated as follows. The performance and feasibility of broadband power line communication for AMI application are investigated in [10-11]. In [12], a wireless architecture is introduced for AMIs based on narrowband IoT, which is considered as a new radio interface to the 3GP. The main disadvantage associated with narrowband IoT is its high power usage. In [13], a wireless architecture was proposed for AMI based on Zigbee technology to provide an interference mitigation mechanism for improving the transmission rate as well as the average delay time. The main problem associated with the proposed architecture is that the Zigbee technology suffers from low transmission range. In [14] the LTE and WiFi are utilized where a hybrid architecture is developed for AMI in which the smart meter uses WiFi and access points use LTE to transfer data. In this study, the problems associated with the above works are solved through the LORA framework.

Another major issue about the data communication and cyber layer in the electrical grids is cyber security. As shown in [3], a successful cyber-attack on AMI can cause severe damages to the system. In the following, recent works about cyber security in electrical grids are investigated. In [1], the LSTM and the concept of prediction intervals are utilized and a hybrid intrusion detection system for data integrity attacks in MGs is proposed. The main disadvantage of the proposed model in [1], is decision making process based on a single observation. In [5], an intrusion detection system based on the sequential probability ratio testing method is proposed to detect identity-based cyber-attacks. In [1] and [5], the cyber and physical layers of the HMG are discussed individually without considering their interaction. In [15], a machine learning-based anomaly detection method based on artificial neural networks (ANN), symbiotic organisms search, and prediction interval was proposed to improve cyber-security in AMIs. In [16], a CADM based on Wavelet transform and deep learning is proposed. Authors in [17] proposed a detection method based on the Hilbert Huang transform and Blockchain for security in enhancement in MGs. The main problem of the above detection methods is that they detect attacks only based on one sample. This is considered as an issue where there is a high level of uncertainties in the system.

While each of the above works has addressed significant topics about the physical and cyber layer of MGs, the research in this area is still in infancy. In order to provide comprehensive research on the operation of HMGs, This paper addresses both cyber and physical layers of HMGs in the same architecture. In the following, some justifications for the proposed methodologies are mentioned. Generally, the mesh network topology has several advantages over traditional star network topology such as better coverage, lower dead zones, lower cost, higher package delivery ratio (PDR), etc. [23]. Also, the mesh network strategy eliminates the need for multiple data gateways in the network. LORA technology has several advantages over other communication technologies. For instance, in contrast to the narrowband IoT networks and cellular-based networks, which have high energy consumption rates; LORA devices have a low energy consumption rate. The LORA devices have a long transmission range of about 2 km in dense urban environments and up to 15 km in rural environments (direct line of sight) where other technologies such as Wi-Fi, Bluetooth, and ZigBee have much lower transmission ranges. The LORA uses CSSM that is robust to multipath and fading issues [19]. In this regard, in this paper, the concept of LORA and mesh network are combined and a mesh network architecture for the AMI based on LORA technology is introduced in which network nodes have peer-to-peer communication. In the proposed architecture, the LORA data frame is designed for smart meters in HMG in presence of WTs, PVs, FCs, and MTs. Due to the high vulnerability of AMIs to cyber-attacks, this paper proposes a CADM based on BLSTM and sequential hypothesis testing (SHT) to detect FDIA on smart meters. The proposed CADM consists of two phases: the prediction phase and the decision-making phase. In the prediction phase, the load consumption is predicted, and in the decision-making phase, the difference between the received load and predicted load is used to decide whether the received data is legitimate or not. The proposed CADM has a major benefit over the CADMs in the literature review. Since there is always some sort of uncertainty associated with the load demand, the proposed SHT, which is a sequential statistical-based decision-making method, is deployed to detect cyber-attacks based on a sequence of samples rather than only one sample. In addition to the cyber layer, this paper presents a mathematical framework for the optimal operation of the HMGs' physical layer as well. Finally, the performance of the proposed framework is evaluated using the modified IEEE 33-bus test system. The test system includes multiple WTs, PVs, MTs, and a fuel cell unit. To highlight the impact of cyber-attacks on the steady-state operation of isolated HMGs, an FDIA is launched against the test system and results are analyzed in detail.

To summarize, the main contributions of this work can be named as below:
- Proposing a cyber-physical architecture for the secured optimal operation of isolated hybrid MGs.
- Proposing a wireless mesh network architecture based on LORA technology for the AMIs.
- Developing a novel deep-learning based CADM utilizing BLSTM and SHT to detect FDIAs in AMIs.
- Simulating an FDIA on the HMG and investigating its effects on the steady-state operation of the system.

The rest of this paper is organized as follows: In section II, the complete details of the proposed CPS architecture including the physical layer, LORA-based network architecture, and CADM is presented. Finally, the simulation results and main conclusion are presented in sections III and IV respectively.

## II. THE PROPOSED CYBER-PHYSICAL ARCHITECTURE

The proposed architecture includes two parts: the physical layer and the cyber layer.

### A. Physical layer: physical operation and optimal scheduling of isolated HMG

In this section, the mathematical model for the optimal operation of the HMGs' physical layer is presented. The cyber layer of the proposed cyber-physical architecture is described in the next sections. The HMG presents an efficient way for the development and operation of electricity grids in the presence of dc energy sources (e.g. PVs and fuel cells) and dc loads as well as legacy ac power systems such that it has the benefits of both ac and dc systems at the same time. In the HMG, ac and dc sub-grids are connected through an ac-dc converter and the system can operate in either grid-connected or isolated mode. In the grid-connected mode, HMG is connected to the upstream grid through the point of common coupling (PCC) and can exchange power based on costs and benefits. In isolated mode, the HMG is isolated from the upstream grid, and loads are supplied using DGs. Due to the absence of the upstream grid, all dispatchable DGs must be scheduled in such a way that the operation cost is minimized and all loads would be supplied. Therefore, the optimal operation of the grid is an important topic that is taken into account in this section. In this section, the physical operation of isolated HMG is formulated as a constrained optimization problem that is solved considering several technical limitations. The cost objective function incorporates the operation cost of HMG as below:

$$Min \quad h(X) = \sum_{t=1}^{N_i}(\sum_{i=1}^{N_d}[u_i^t P_{Gi}^t C_{Gi}^t + S_{Gi}^{on} \max\{0, u_i^t - u_i^{t-1}\} + \quad (1)$$

$$+ S_{Gi}^{off} \max\{0, u_i^{t-1} - u_i^t\}])$$

where the variable $X$ in (1) presents the vector of decision variables:

$$X = [P_g, U_g]_{1 \times (2 \times N_d \times N_i)} \quad ; \quad \forall t \in N_i$$
$$P_g^t = [P_{G1}^t, P_{G2}^t, ..., P_{GN_d}^t]; U_g^t = [u_1^t, u_2^t, ..., u_{N_d}^t] \quad u_k^t \in \{0,1\} \quad (2)$$

The cost objective function in (1) must be optimized considering several technical and practical constraints as below:
- Generation-demand balance in dc sub-grid:

$$\sum_{i=1}^{N_{d-DC}} P_{Gi}^t + P_{conv}^t = \sum_{k=1}^{N_{Load-DC}} P_{Load-DC}^t + P_{Loss-DC} \quad (3)$$

- Active and reactive powers balance in ac sub-grid:

$$P_j^{inj,t} = \sum_{n=1}^{N_b} V_j^t V_n^t Y_{jn} \cos(\theta_{jn} + \delta_j - \delta_n) \quad (4)$$

$$Q_j^{inj,t} = \sum_{n=1}^{N_b} V_j^t V_n^t Y_{jn} \sin(\theta_{jn} + \delta_j - \delta_n) \quad (5)$$

In this study, the forward-backward method is utilized to preserve the above equations.

- Capacity limitation of converter and DGs:

$$P_{Gi,\min}^t \leq P_{Gi}^t \leq P_{Gi,\max}^t$$
$$P_{conv,\min}^t \leq P_{conv}^t \leq P_{conv,\max}^t \quad (6)$$

- Spinning reserve:

$$\sum_{i=1}^{N_d} u_i^t P_{Gi,\max}^t \geq \sum_{k=1}^{N_{Load}} P_{Load,k}^t + P_{loss}^t + \text{Re } s^t \quad (7)$$

- Feeder capacity limit:

$$|P_i^{Line,t}| \leq P_{i,\max}^{Line} \quad (8)$$

- Bus voltage limit:

$$V_m^{\min} \leq V_m^t \leq V_m^{\max} \quad (9)$$

- Ramp-rate constraint:

$$|P_{Gi}^t - P_{Gi}^{t-1}| < UR_i, DR_i \quad (10)$$

### B. Cyber layer: LORA based network architecture and deep-learning based cyber-attack detection model

While the previous section presented the physical operation of the HMG, this section is devoted to the cyber layer of the proposed architecture. In this section, the LORA-based network architecture and deep-learning based cyber-attack detection model are presented.
3

*B.1. LORA based architecture for AMI*

It is worth noting that only the Theoretical details of the proposed LORA-based network are presented in this study . The numerical simulations and practical implementation of the proposed LORA framework are considered as future work. Low power wide area network (LPWAN) is a class of wireless communication that is designed for transmitting small data packets over long distances. There are several technologies in the LPWAN area such as SigFox, Narrowband IoT, and LORA. The main advantage of LPWANs over the other communication technologies (e.g. ZigBee, Bluetooth, Wi-Fi, LTE, etc.) is the low power usage and long transmission range. LORA is a LPWAN technology that is designed and developed by SemTech and uses the chirp spread spectrum modulation (CSSM) technique to transfer small data packages (0.3 kbps to 5.5 kbps) over long distances [22].

A LORA node is often battery-powered and consists of two parts: 1) a radio module with an antenna, and 2) A microprocessor to process sensor data. The minimum received signal strength (RSS) value for LORA is -120 decibel milliwatts (dBm), the maximum antenna gain is +2.15 decibel relative to isotrope (dBi), the maximum link budget is 157 dB, and the maximum transmission power for uplink and downlink are 14 dBm and 27dBm, respectively. Also, LORA receivers have sensitivity level down to -148 dBm [18]. LORA operates in the unlicensed industrial, scientific, and medical (ISM) radio band that is available worldwide. The ISM radio band regulation is specified based on the region's regulation. For instance, in Europe, the ISM band ranges from 863MHz to 870MHz. Since the ISM radio band is a public band, anyone is allowed to use these frequencies and no license fee is required. In the proposed architecture, the smart meters communicate with the central control every hour, and the class B LORA devices [20] are utilized.

The network's initial setup and routing process are based on the protocol described in [23]. In the proposed network, each smart meter in the network is equipped with a microprocessor (ARM M0 microprocessor) to process commands and also a Semtech LORA transceiver (SX1278 LORA RF transceiver) to send and receive signals. There is a gateway in each region that collects the data of smart meters in that region and sends them to the central control for monitoring and processing purposes. In [23], the mesh network setup and routing protocol is practically tested using a LORA wireless mesh network including 19 nodes and one gateway to monitor a 600m×800m university campus. In that experiment, the SF=12, bandwidth=125 kHz, and also the maximum query latency and the gateway query interval are considered as one minute. The results showed that the LORA mesh network architecture can reach around 89% packet delivery ratio where the LORA star network topology under the same setting achieved only 58.7%. This shows the superiority of the mesh network topology over the traditional one-hop star network topology.

*B.1.1. LORA data frame: the case of smart meter*

In this section, the LORA data frame is described in detail. Also, a novel design of LORA data packet for smart meters in HMGs is proposed. Generally, a LORA data packet consists of three layers: the physical layer (PHY), the MAC layer, and the application layer [18]. Fig. 1 shows the LORA data frame structure.

- Physical layer: The physical layer data frame starts with a preamble. The preamble is responsible to synchronize the receiver and transmitter and also it indicates the packet modulation scheme. The physical header (PHDR), which is 20 bits, indicates the data length, error correction coding rate, whether the PHY cyclic redundancy check (PHY_CRC) is included at the end of the data packet, and also includes a 4-bit PHDR_CRC. Note that there are two types of PHDRs: explicit and implicit. The implicit PHDR is utilized for downlink messages and doesn't include PHY_CRC.

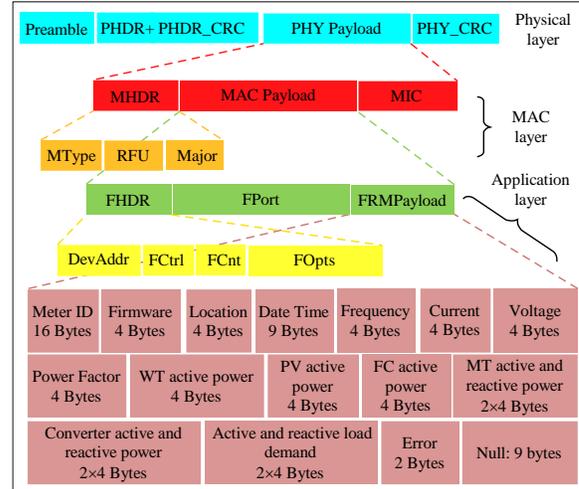

Fig. 1. LORA data frame in the context of smart meter

- MAC layer (PHY payload): the LORA data frame in the MAC layer consists of three parts: MAC header (MHDR), MAC payload, and message integrity code (MIC). The MIC is a 4-byte hash that ensures that the data is not changed in the path. If the received MIC doesn't match the expected hash of data, the receiver drops the received packet. Note that in a join-accept frame, the MIC is encrypted with the MAC payload and is not a separate field. The MHDR, which is 1 byte, includes 2 bits to specify the major version data frame (Mjor) that is utilized to encode the data, 3 bits to indicate message type (MType), and 3 bits to reserve for future usage (RFU). It is worth noting that there are 8 types of messages in LORA (e.g. join request (MType=000), proprietary(MType=111), etc.).

- Application layer (MAC payload): The application layer, consists of three parts: frame header (FHDR), port field (FPort), and frame payload (FRMPayload). The FHDR contains 4 bytes for the short device address of the end-device (DevAddr), 2 bytes for the frame counter (FCnt), 1 byte for the frame control (FCtrl), and up to 15 bytes for the frame options (FOpts). The FPort contains 8 bits that can present 256 decimal values (i.e. from 0 to 255). The FPort value 0 indicates the MAC command; FPort values from 1 to 223 are application-specific and are available to the application layer. The FPort value 224 presents the LORA MAC layer test protocol, and the FPort values higher than 224 are discarded. The FOpts are encrypted based on the network session encryption key that is utilized to encrypt MAC payload and calculate MIC. The FRMPayload is encrypted based on the generic algorithm encryption scheme in IEEE 802.15.4 with a key length of 128 bits, which should be encrypted before MIC calculation. The maximum FRMPayload is specified based on worldwide region and the data rate parameter as described in [21].

Fig. 1 shows the LORA FRMpayload structure designed for smart meters in HMGs in presence of WTs, PVs, MTs, and FCs. In the proposed data frame, various parameters such as frequency, current, active power, reactive power, power factor, timestamp, etc. are considered. The structure in Fig. 1 is applied to all smart meters in the system. Note that if a node doesn't include a specific component (e.g. WT, PV, etc.), the component's corresponding part in the data packet becomes all one. For instance, if a specific node doesn't include WT, the 4-byte WT active power part in the data frame should present the decimal value of $2^{32}-1$ (i.e. $[1, 1, …, 1]_{32}$). Therefore, the decimal value of $2^{32}-1$ in the data packet related to a specific component, presents the absence of that specific component in the node. The structure in Fig. 1 is designed considering spreading factors (SF) 6, 7, and 8 where the data packet contains 96 bytes of data [23]. In the case of SFs 9, 10, 11, and 12, which have a lower maximum FRMPayload size, the data is divided into two 51-byte packets. In the network server, these two 51-byte data packets are identified using a two-byte packet identifier and combined to form the 96-byte data packet.

*B.2. Deep learning-based cyber-attack detection model*

In this section, a deep learning-based CADM based on BLSTM and SHT is proposed to detect FDIAs on the measured load demand data in smart grids. This model is implemented and used in central control. When the data related to a load meter is received at the central control, it goes through this algorithm to check whether the data sending meter is under attack or not. This process is constantly happening in the system for all meters in the system. The proposed CADM includes two phases: the prediction phase and the decision-making phase. After data is received in the central control, in the prediction phase, the received data will be predicted using BLSTM also. Then, in the detection phase, the difference between the measured and predicted data will be computed and used to decide whether the corresponding meter is under attack or not. Since there is uncertainty associated with the load demand in practice, it is not reasonable to make decisions about the integrity of measured load based on only one piece of evidence (i.e. one received data). To this end, the SHT, which is a sequential-statistical decision-making method, is utilized to make decisions based on the sequence of evidence. This means that the algorithm views several consecutive data to decide whether the transmitter meter is being attacked and this process of sequential observation is done by SHT.

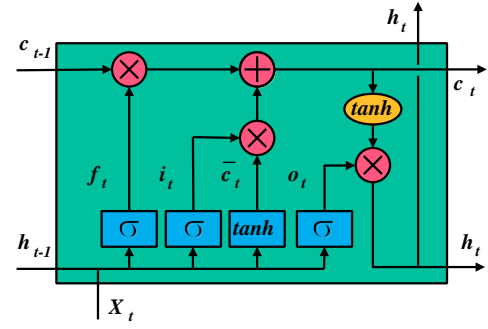

Fig. 2 structure of LSTM cell

*B.2.1. Prediction phase: Load prediction using bidirectional LSTM*

In this section, the mathematical of the BLSTM is provided. The BLSTM algorithm is utilized to forecast the received measured load demand data. The BLSTM is trained offline using historical data and the trained model is used to detect attacks. In order to overcome the gradient vanishing problem in the conventional recurrent neural networks, the concept of long short-term memory (LSTM) is introduced [24]. The main beneficial characteristic of LSTM is its cell state and three cell gates called input gate, forget gate, and output gate. Cell gates are responsible to maintain and adjust the cell state. The forget gate specifies the information that should be removed from the cell state, the input gate defines which information should be added to the cell state, and the output gate specifies which information from the cell state should be used in the output. Fig. 2 shows the structure of the LSTM cell.

Given an input sequence $x = (x_1, x_2, …, x_T)$, an LSTM cell computes the hidden state sequence $h = (h_1, h_2, …, h_T)$ and output sequence $y = (y_1, y_2, …, y_T)$ as below [26]:

$$h_t = H(W_{x,h}x_t + W_{h,h}h_{t-1} + b_h) \tag{11}$$

$$y_t = W_{h,y}h_t + b_y \tag{12}$$

The function H is implemented as follows:

$$i_t = \partial(W_{i,x}x_t + W_{i,h}h_{t-1} + b_i) \tag{13}$$

$$f_t = \partial(W_{f,x}x_t + W_{f,h}h_{t-1} + b_f) \tag{14}$$

$$\overline{c}_t = \tanh(W_{\overline{c},x}x_t + W_{\overline{c},x}h_{t-1} + b_{\overline{c}}) \tag{15}$$

$$o_t = \partial(W_{o,x}x_t + W_{o,h}h_{t-1} + b_o) \tag{15}$$

$$C_t = f_t \cdot c_{t-1} + i_t \cdot \overline{c}_t \tag{17}$$

$$h_t = o_t \cdot \tanh(c_t) \tag{18}$$

Furthermore, the LSTM network's weights and biases are adjusted based on the backpropagation technique [25] during the training process.

One drawback associated with LSTMs is that they can only make use of previous information. The concept of BLSTM, which is introduced in [27], makes it possible to use both previous and future contexts. The schematic structure of the BLSTM network has been illustrated in Fig. 3. In BLSTMs, two layers operate in reversed time step directions in such a way that the forward hidden sequence ($\vec{h}$), backward hidden sequence ($\overleftarrow{h}$), and output sequence ($y$) are generated by iterating the forward layer from $t=1$ to $t=T$, backward layer from $t=T$ to $t=1$, and computing the output as follows [26]:

$$\vec{h}_t = H(W_{x,\vec{h}}x_t + W_{\vec{h},\vec{h}}\vec{h}_{t-1} + b_{\vec{h}}) \qquad (19)$$

$$\overleftarrow{h}_t = H(W_{x,\overleftarrow{h}}x_t + W_{\overleftarrow{h},\overleftarrow{h}}\overleftarrow{h}_{t-1} + b_{\overleftarrow{h}}) \qquad (20)$$

$$y_t = W_{\vec{h},y}\vec{h}_t + W_{\overleftarrow{h},y}\overleftarrow{h}_t + b_y \qquad (21)$$

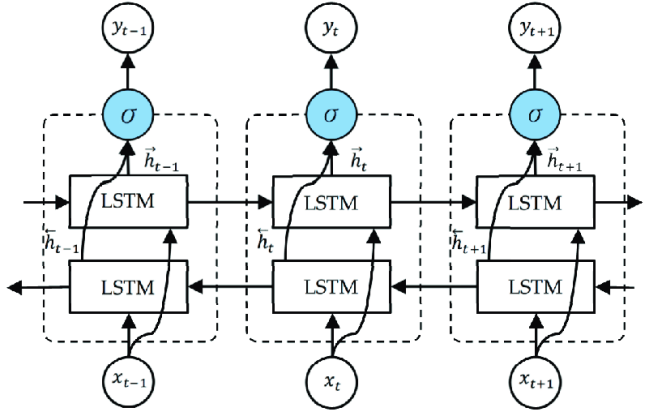

Fig. 3 Schematic structure of the BLSTM network

*B.2.2. Detection phase: Decision making using SHT*

When a measured load data is received in the HMG central control (HMGCC), the load value is compared with its corresponding predicted value (prediction phase), and a binary sample is generated based on their difference as follows:

$$e_{l,i} = |P_{l,m}^i - P_{l,f}^i| \qquad (22)$$

$$S_l^i = \begin{cases} 0 & 0 \le \dfrac{e_l^i}{P_{l,f}^i} \le LE^i \\ 1 & LE^i < \dfrac{e_l^i}{P_{l,f}^i} \le UE^i \end{cases} \qquad (23)$$

The value of $LE^i$ and $UE^i$ are determined by system operators based on historical data in such a way that the forecast error ($e_{l,i}$) lower than $LE^i \times P_{l,f}^i$ is common in the system, errors in the range of ($LE^i \times P_{l,f}^i$, $UE^i \times P_{l,f}^i$] are rare but happens from time to time, and forecast error greater than $UE^i \times P_{l,f}^i$ never happens in the system. Note that data with forecast error greater than $UE^i \times P_{l,f}^i$ is directly considered as an attack with only one sample. The $S_l^i$ is used as the input of SHT and the output is whether the sequence of input samples is legitimate or not. SHT, which is a sequential-statistical decision-making method, is a one-dimensional random walk with two thresholds that starts from a point between thresholds and move toward one of them [28]. In SHT, two hypotheses and two thresholds are considered in such a way that the upper threshold ($U^i$) and lower threshold ($L^i$) are associated with the alternative hypothesis ($H_1$) and the null hypothesis ($H_0$). Hypotheses are defined as below:

- $H_1$: smart metering device is under attack.
- $H_0$: smart metering device is not under attack.

In SHT, the decision is always made based on a finite number of samples. For a finite number $n$, ($S_{l,1}^i, S_{l,2}^i, ..., S_{l,n}^i$) is defined as a sample set that contains the sequence of $n$ input samples. According to [28], the probability ratio of a sample set of size $n$ is computed as follows:

$$PR_T^i = \ln(\prod_{k=1}^n \dfrac{f(S_{l,k}^i|H_1)}{f(S_{l,k}^i|H_0)}) = \sum_{k=1}^n PR_k^i = \qquad (24)$$

$$= m \times \ln(\dfrac{P_1^i}{P_0^i}) + (n-m) \times \ln(\dfrac{1-P_1^i}{1-P_0^i})$$

Where:

$$\begin{cases} P_0^i = PR_k^i\left(S_{l,k}^i = 1 \mid H_0\right) \\ P_1^i = PR_k^i\left(S_{l,k}^i = 1 \mid H_1\right) \end{cases} \qquad (25)$$

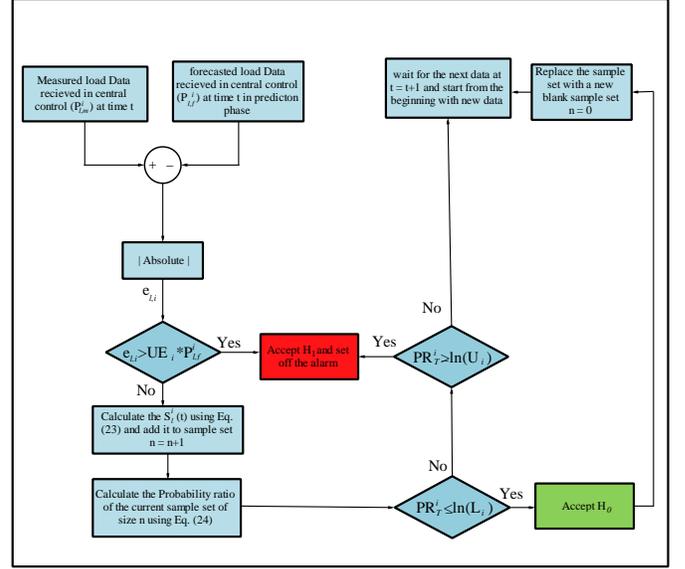

Fig. 4 flowchart diagram of the proposed CADM.

Note that input samples are independent and identically distributed. At each step of the process, when a new measured load data is received, the input sample ($S_l^i$) is calculated and added to the current sample set and the probability ratio of the sample set is calculated. If $PR_T^i \le Ln(L^i)$ the $H_0$ is accepted, if $Ln(U^i) \le PR_T^i$, the $H_1$ is accepted, and if $Ln(L^i) < PR_T^i < Ln(U^i)$, no decision is made and the process waits for the next input sample. After a decision is made (i.e. acceptance of $H_1$ or $H_0$), the sample set is cleared and a new blank sample set is generated for the next samples (n=0). Flowchart diagram of the proposed CADM is presented in Fig. 4. According to [28], the thresholds and expected value of number of samples required by the SHT to reach a decision ($N^i$) are calculated as follows:

$$E(N^i \mid H_0) = \dfrac{(1-\alpha^i) \times \ln(L^i) + \alpha^i \times \ln(U^i)}{P_0^i \times \ln(\dfrac{P_1^i}{P_0^i}) + (1-P_0^i)\ln(\dfrac{1-p_1^i}{1-p_0^i})} \qquad (26)$$

$$E(N^i \mid H_1) = \dfrac{\beta^i \times \ln(L^i) + (1-\beta^i) \times \ln(U^i)}{P_1^i \times \ln(\dfrac{P_1^i}{P_0^i}) + (1-P_1^i)\ln(\dfrac{1-p_1^i}{1-p_0^i})} \qquad (27)$$

$$U^i = ((1-\beta^i)/\alpha^i) \qquad (28)$$

$$L^i = (\beta^i / (1-\alpha^i)) \qquad (29)$$

TABLE I
CHARACTERISTICS OF DGS AND AC-DC CONVERTER

| Type | Min Power (kW) | Max Power (kW) | Bid ($/kWh) | Startup/ shutdown cost ($) | Ramp Up/Down Rate | Bus Number |
|---|---|---|---|---|---|---|
| MT 2 | 100 | 1300 | 0.475 | 70 | 185 | 12 |
| MT 3 | 90 | 1100 | 0.475 | 75 | 150 | 25 |
| WT 2 | 0 | 550 | 1.073 | 0 | - | 30 |
| WT 3 | 0 | 450 | 1.073 | 0 | - | 21 |
| PV 2 | 0 | 400 | 2.584 | 0 | - | 16 |
| AC-DC converter | -1000 | 1000 | - | - | - | 18 |
| FC | 50 | 700 | 0.494 | 38.5 | 110 | DC MG |
| WT 1 | 0 | 200 | 1.073 | 0 | - | DC MG |
| MT 1 | 35 | 300 | 0.48 | 60 | 60 | DC MG |
| PV 1 | 0 | 250 | 2.584 | 0 | - | DC MG |

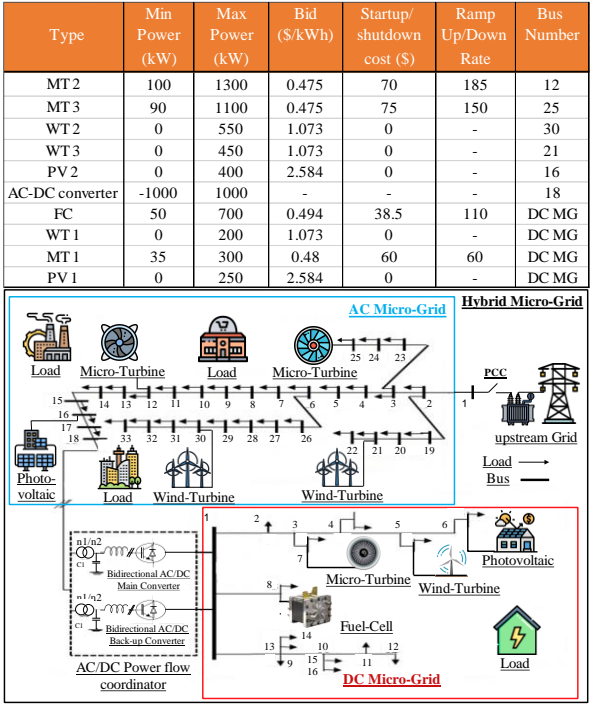

Fig. 5. Schematic illustration of the test system

## III. SIMULATION RESULTS

In this section, the modified IEEE 33-bus test system is utilized to evaluate the performance of the proposed architecture. The test system includes three WTs, two PVs, three MTs, one FC, and a dc sub-grid which is connected to the ac grid through bidirectional ac-dc converters at bus 18. The schematic illustration of the test system is presented in Fig. 5. The characteristics of DGs and converter can be found in Table I. Ac MG and dc MG have 12.66 kV and 1 kV voltage levels, RESs (i.e. WTs and PVs) are considered as non-dispatchable units, and active power loss in only dc sub-grid is neglected. Complete information of WTs and PVs' generation pattern, ac MG load factor, and dc MG load demand are presented in Table II. Note that the hourly load of each bus is obtained by multiplying the load factor by the maximum load of that bus. The results are provided for 24 hours considering one-hour intervals.

Additionally, the hourly load data of the utility in the city of Johor in Malaysia in 2009 and 2010 [29] is used to examine the performance of the proposed deep-learning-based CADM on a practical test case. Finally, in order to show the interaction between cyber and physical layers of the system and also to investigate the impact of cyber-attacks on the optimal operation of HMGs, an FDIA with 70% severity is launched against the test system and results are analyzed in detail. Note that FDIA with 70% severity means that the adversary changes the measured data by 70% of its actual value. In this section, the dragonfly optimization [30] is used to solve the grid's operation problem that is formulated as a single objective optimization problem (see section II). As mentioned before, the optimal scheduling is carried out for 24 hours considering one-hour intervals. Table III shows the optimal output power of DGs and ac-dc converter. Note that negative values of the ac-dc converter present power flow from the dc sub-grid to the ac grid. Since the power generation capacity in the dc sub-grid is more than the consumption, the dc sub-grid injects power into the ac grid during the day. As can be seen from Table III, due to the absence of an upstream grid, all DGs operate near their maximum generation capacity at peak load hours. Also, since the generation capacity in the dc sub-grid is greater than demand, the power direction is toward the ac sub-grid in all hours (see negative values for converter in Table III). According to Table V, the maximum voltage deviation constraint (i.e. 0.1 pu) is satisfied.

Table II. hourly dc MG load demand, PVs normalized generation pattern, WTs normalized generation pattern, ac MG load factor [5].

| Hour | ac MG load factor | dc MG load demand | WT normalized output | PV normalized output |
|---|---|---|---|---|
| 1 | 0.6 | 156 | 0.119 | 0 |
| 2 | 0.65 | 150 | 0.119 | 0 |
| 3 | 0.59 | 150 | 0.089 | 0 |
| 4 | 0.62 | 153 | 0.15 | 0 |
| 5 | 0.7 | 168 | 0.204 | 0 |
| 6 | 0.698 | 174 | 0.18 | 0 |
| 7 | 0.71 | 210 | 0.24 | 0.109 |
| 8 | 0.79 | 225 | 0.26 | 0.25 |
| 9 | 0.86 | 228 | 0.26 | 0.34 |
| 10 | 0.9 | 240 | 0.3 | 0.39 |
| 11 | 0.98 | 234 | 0.29 | 0.468 |
| 12 | 1 | 222 | 0.31 | 0.47 |
| 13 | 0.99 | 216 | 0.29 | 0.461 |
| 14 | 1 | 216 | 0.27 | 0.5 |
| 15 | 0.97 | 228 | 0.285 | 0.47 |
| 16 | 0.958 | 240 | 0.298 | 0.35 |
| 17 | 0.935 | 255 | 0.33 | 0.26 |
| 18 | 0.86 | 264 | 0.35 | 0.19 |
| 19 | 0.88 | 270 | 0.4 | 0.04 |
| 20 | 0.91 | 261 | 0.45 | 0 |
| 21 | 0.927 | 234 | 0.42 | 0 |
| 22 | 0.887 | 213 | 0.39 | 0 |
| 23 | 0.78 | 195 | 0.36 | 0 |
| 24 | 0.7 | 168 | 0.22 | 0 |

### A. Hybrid isolated MG under FDIA

In this section, we simulate FDIA against the measured load consumption of buses 7, 8, 20, 21, 24, 25, 29, 30, 31 & 32 that are the most-loaded buses in the HMG that contain a total load of 2100 kW. Note that this attack is launched against the test system to show the effect of worst-case undetected FDIA scenarios against the test system, although the proposed CADM can detect cyber-attacks with significantly lower intensities as well. The cyber-attack is launched at the twelfth hour of the day and the load demand of the buses is reduced by 70% of their actual demand. The moment the attack is launched, HMGCC observes 1470 kW unpredicted excess power in the system. At this moment, HMGCC tends to decrease the power generation in the system, but since the HMGCC is in isolated mode and the maximum ramp-rate of DGs in the system is limited (i.e. 505 kW), if all DGs reduce their generation as much as they can, there will be still 920.5 kW excess power in the system. At this point, HMGCC has no choice but to send an emergency shutdown command to at least one DG. Therefore, HMGCC sends an emergency shutdown command to MT 3, which has the closest output power to 1450 kW (i.e. 1300 kW), to strike a balance between generation and consumption in the system. Note that all of the above decisions were made based on false data originating from hacking the meters. The moment MT 3 turns off, the real balance in the system is lost and frequency deviates from the allowed range. At this point, the only way that can keep frequency in the range is to cut off some loads. After the shutdown-startup period of the MT 3 passed, it starts generating again. However, due to its ramp-rate limitation, its output power should be increased gradually.

TABLE III
OPTIMAL SCHEDULING OF THE HMG

| Hour | PV1 | WT1 | FC | MT1 | AC-DC Converter | MT2 | MT3 | WT2 | WT3 | PV2 |
|---|---|---|---|---|---|---|---|---|---|---|
| 1 | 0 | 23.8 | 438.6 | 133.2 | -439.58 | 564.1 | 1098.3 | 65.45 | 53.55 | 0 |
| 2 | 0 | 23.8 | 545.3 | 191.6 | -610.7 | 650.1 | 1037 | 65.45 | 53.55 | 0 |
| 3 | 0 | 17.8 | 482.5 | 251 | -601.31 | 552.4 | 947.85 | 48.95 | 40.05 | 0 |
| 4 | 0 | 30 | 440.1 | 214.7 | -531.8 | 585.8 | 1031 | 82.5 | 67.5 | 0 |
| 5 | 0 | 40.8 | 549.9 | 274.5 | -697.17 | 604.7 | 1100 | 112.2 | 91.8 | 0 |
| 6 | 0 | 36 | 572.2 | 294.9 | -729.11 | 591.5 | 1100 | 99 | 81 | 0 |
| 7 | 27.25 | 48 | 469.8 | 242 | -577.06 | 678.4 | 1100 | 132 | 108 | 43.6 |
| 8 | 62.5 | 52 | 579.4 | 298.5 | -767.39 | 743.8 | 1084.7 | 143 | 117 | 100 |
| 9 | 85 | 52 | 635.8 | 299.8 | -844.57 | 928.8 | 1066 | 143 | 117 | 136 |
| 10 | 97.5 | 60 | 590.9 | 240 | -748.43 | 1113 | 1068.6 | 165 | 135 | 156 |
| 11 | 117 | 58 | 681.2 | 211.7 | -833.94 | 1298 | 1100 | 159.5 | 130.5 | 187.2 |
| 12 | 117.5 | 62 | 700 | 235.4 | -892.88 | 1300 | 1100 | 170.5 | 139.5 | 188 |
| 13 | 115.3 | 58 | 684 | 236 | -877.23 | 1300 | 1100 | 159.5 | 130.5 | 184.4 |
| 14 | 125 | 54 | 669.1 | 295 | -927.15 | 1300 | 1100 | 148.5 | 121.5 | 200 |
| 15 | 117.5 | 57 | 675.3 | 295.1 | -916.92 | 1300 | 992.28 | 156.8 | 128.3 | 188 |
| 16 | 87.5 | 59.6 | 606.1 | 282.8 | -796 | 1300 | 1084.2 | 163.9 | 134.1 | 140 |
| 17 | 65 | 66 | 604.7 | 223.4 | -704.11 | 1300 | 1080.7 | 181.5 | 148.5 | 104 |
| 18 | 47.5 | 70 | 576.7 | 189.1 | -619.35 | 1153 | 1021.8 | 192.5 | 157.5 | 76 |
| 19 | 10 | 80 | 685.2 | 182.5 | -687.71 | 1300 | 899.47 | 220 | 180 | 16 |
| 20 | 0 | 90 | 629.5 | 237.9 | -696.49 | 1300 | 967.5 | 247.5 | 202.5 | 0 |
| 21 | 0 | 84 | 584.1 | 220.7 | -654.81 | 1300 | 1100 | 231 | 189 | 0 |
| 22 | 0 | 78 | 559.5 | 161.9 | -586.42 | 1274 | 1068.7 | 214.5 | 175.5 | 0 |
| 23 | 0 | 72 | 451.2 | 101.9 | -430.13 | 1089 | 1021.7 | 198 | 162 | 0 |
| 24 | 0 | 44 | 341.6 | 43.09 | -260.72 | 1020 | 1093.9 | 121 | 99 | 0 |

TABLE IV
OUTPUT POWER OF DGS AND CONVERTER UNDER FDIA

| Hour | PV1 | WT1 | Fuel Cell | MT1 | AC-DC Converter | MT2 | MT3 | WT2 | WT3 | PV2 | Load Shedding |
|---|---|---|---|---|---|---|---|---|---|---|---|
| 1 | 0 | 23.8 | 438.6 | 133.2 | -439.58 | 564.1 | 1098.3 | 65.45 | 53.55 | 0 | 0 |
| 2 | 0 | 23.8 | 545.3 | 191.6 | -610.7 | 650.1 | 1037 | 65.45 | 53.55 | 0 | 0 |
| 3 | 0 | 17.8 | 482.5 | 251 | -601.31 | 552.4 | 947.85 | 48.95 | 40.05 | 0 | 0 |
| 4 | 0 | 30 | 440.1 | 214.7 | -531.8 | 585.8 | 1031 | 82.5 | 67.5 | 0 | 0 |
| 5 | 0 | 40.8 | 549.9 | 274.5 | -697.17 | 604.7 | 1100 | 112.2 | 91.8 | 0 | 0 |
| 6 | 0 | 36 | 572.2 | 294.9 | -729.11 | 591.5 | 1100 | 99 | 81 | 0 | 0 |
| 7 | 27.25 | 48 | 469.8 | 242 | -577.06 | 678.4 | 1100 | 132 | 108 | 43.6 | 0 |
| 8 | 62.5 | 52 | 579.4 | 298.5 | -767.39 | 743.8 | 1084.7 | 143 | 117 | 100 | 0 |
| 9 | 85 | 52 | 635.8 | 299.8 | -844.57 | 928.8 | 1066 | 143 | 117 | 136 | 0 |
| 10 | 97.5 | 60 | 590.9 | 240 | -748.43 | 1113 | 1068.6 | 165 | 135 | 156 | 0 |
| 11 | 117 | 58 | 681.2 | 211.7 | -833.94 | 1298 | 1100 | 159.5 | 130.5 | 187.2 | 0 |
| 12 | 117.5 | 62 | 700 | 271.7 | -929.2 | 100 | 1100 | 170.5 | 139.5 | 188 | 1115.95 |
| 13 | 115.3 | 58 | 700 | 300 | -957.25 | 285 | 1100 | 159.5 | 130.5 | 184.4 | 892.897 |
| 14 | 125 | 54 | 700 | 300 | -963 | 470 | 1100 | 148.5 | 121.5 | 200 | 750.909 |
| 15 | 117.5 | 57 | 700 | 300 | -946.5 | 655 | 1100 | 156.8 | 128.3 | 188 | 470.881 |
| 16 | 87.5 | 59.6 | 700 | 300 | -907.1 | 840 | 1100 | 163.9 | 134.1 | 140 | 316.25 |
| 17 | 65 | 66 | 700 | 300 | -876 | 1025 | 1100 | 181.5 | 148.5 | 104 | 84.4836 |
| 18 | 47.5 | 70 | 590 | 240 | -683.5 | 1091 | 1021.8 | 192.5 | 157.5 | 76 | |
| 19 | 10 | 80 | 685.2 | 182.5 | -687.71 | 1276 | 922.22 | 220 | 180 | 16 | 0 |
| 20 | 0 | 90 | 629.5 | 237.9 | -696.49 | 1300 | 967.5 | 247.5 | 202.5 | 0 | 0 |
| 21 | 0 | 84 | 584.1 | 220.7 | -654.81 | 1300 | 1100 | 231 | 189 | 0 | 0 |
| 22 | 0 | 78 | 559.5 | 161.9 | -586.42 | 1274 | 1068.7 | 214.5 | 175.5 | 0 | 0 |
| 23 | 0 | 72 | 451.2 | 101.9 | -430.13 | 1089 | 1021.7 | 198 | 162 | 0 | 0 |
| 24 | 0 | 44 | 341.6 | 43.09 | -260.72 | 1020 | 1093.9 | 121 | 99 | 0 | 0 |

Table IV shows the hourly load shedding and output power of DGs and converter under cyber-attack. As can be seen from this table, a successful attack can affect the grid for several hours and cause widespread load shedding. According to Table V, although the grid's power loss decreased in the attack scenario, its operation cost has significantly increased which is due to the energy not supplied penalty costs. It is worth noting that the dynamic effect of FDIA on the operation of the HMG as well as shutdown-startup period of the MT 3 are neglected and considered as future work, and also the energy not supplied penalty cost is considered as the maximum energy market price in grid-connected mode 4 $/kWh [5]. However, the energy not supplied penalty cost can be different for different systems with different policies.

### B. Deep learning-based CADM performance evaluation

In this part, the practical hourly load data of the power supply company of Johor city in Malaysia is used to evaluate the performance of the proposed deep learning-based CADM. In this regard, first, the BLSTM network is trained using 15 months of hourly load data generated from 01/01/2009 to 14/03/2010. Then, 3.5 months of data from 15/03/2010 to 02/07/2010 is used to test the trained network. Prediction models are implemented using Tensorflow.Keras library in python 3.8. It is worth noting that in the training process, a 14-hour time lag window, as well as the data samples' day of the week, are considered and the test result of the BLSTM is compared with ANN and regular LSTM under the same setting.

TABLE V
TOTAL OPERATION COST, POWER LOSS, AND MAXIMUM VOLTAGE DEVIATION WITH AND WITHOUT FDIA

| Operation Mode | Case | Power Loss (kW) | Total Cost ($) | Maximum Voltage Deviation (pu) |
|---|---|---|---|---|
| Islanded mode | No attack | 2524 | 48768 | 0.068 |
| | %70 Attack | 2339 | 61487 | 0.051 |

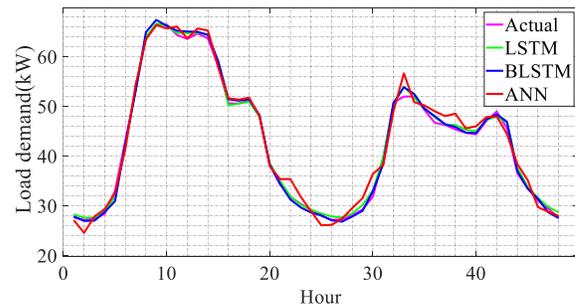

Fig. 6. actual load demand vs predicted load using BLSTM, ANN, and LSTM

The BLSTM and LSTM networks consist of two 128-cell layers, one dropout layer with a 0.3 rate, and a dense output layer with a linear activation function. The networks are trained considering mean square error (MSE) as loss function, Adam as the optimizer, and 150 epochs. Fig. 6 shows the actual load demand vs predicted load using different models for a 48-hour interval from 19/03/2010-00:00 to 21/03/2010-23:00. The high accuracy and accurate tracking capability of the proposed deep learning based method can be seen at sharp points. Also, Table VI compares the root mean square error (RMSE), mean absolute percentage error (MAPE), and mean absolute error (MAE) of models [31]. As can be seen from Table VI, the BLSTM outperforms the ANN and LSTM.

Now that the network is trained, it is time to simulate the FDIA and evaluate the performance of the proposed CADM.

To this end, a scenario is considered in which an adversary penetrates the system at 19/03/2010-12:00 and increases the load demand for five hours. According to the training data, which is used as the system's historical data, and prediction results of the training data using BLSTM, $UE^i$=24.59, and $LE^i$=0.08. The training data set, contains 10511 load consumption samples where 99 samples have forecast error in the range of ($LE^i \times P^i_{l,f}$, $UE^i \times P^i_{l,f}$], therefore $P_0$=0.0094. Since there is no information about cyber-attacks and attacked load consumption samples in the data set, the value of $P_1$ is unknown. In this work, the value of $P_0$ is considered as $P_1$=0.99, which is a reasonable value in practice [5]. Also the false positive and false negative values are considered as $\alpha^i$=0.001 and $\beta^i$=0.002. The attacked load consumption samples are generated using a uniform random distribution such that their $e_{l,i} \in (LE^i \times P^i_{l,f}, UE^i \times P^i_{l,f}]$. Note that input load samples with $e_{l,i} > UE^i \times P^i_{l,f}$ is directly considered as attacking without sequential testing. Table VII shows the performance of the proposed CADM. Red numbers in that table show the false data which has a considerable difference with actual load demand. The no-attack decision in Table VII means that the CADM cannot certainly decide whether the sequence of received data is legitimate or not and needs more evidence. As can be seen from that table, the proposed CADM could detect the attack with only two samples. The moment the attack was detected; system operators can take proper actions based on their instructions.

TABLE VI
MAPE, MAE, AND RMSE OF DIFFERENT PREDICTION MODELS

|  | MAPE(%) | MAE(W) | RMSE(W) |
| --- | --- | --- | --- |
| LSTM | 2.1928 | 931.63 | 1361.8 |
| ANN | 4.1529 | 1676.2 | 2258.8 |
| BLSTM | 2.004 | 866.71 | 1322.4 |

TABLE VII
THE PERFORMANCE OF THE PROPOSED CADM WITH FDIA

| Hour | Actual load demand(W) | Received load Demand(W) | Probability ratio vs thresholds | Decision |
| --- | --- | --- | --- | --- |
| 8:00 | 66364 | 66364 | $Ln(L^i)<-4.59<ln(U^i)$ | No decision |
| 9:00 | 66454 | 66454 | $-9.19<ln(L^i)$ | No attack |
| 10:00 | 64382 | 64382 | $Ln(L^i)<-4.59<ln(U^i)$ | No decision |
| 11:00 | 63589 | 63589 | $-9.19<ln(L^i)$ | No attack |
| 12:00 | 64724 | 72641 | $Ln(L^i)<+4.65<ln(U^i)$ | No decision |
| 13:00 | 63692 | 74133 | $Ln(L^i)<+9.31<ln(U^i)$ | Attack |
| 14:00 | 58141 | 69535 | - |  |
| 15:00 | 50550 | 62065 | - |  |
| 16:00 | 50584 | 62680 | - |  |

IV. CONCLUSION

Due to the high importance of the cyber and physical operation of MGs, this paper proposed a cyber-physical architecture for isolated HMGs. In order to investigate the operation of isolated HMGs under cyber-attacks, an FDIA was launched against the test system. The results showed that a successful attack can cause widespread damage to the system such that an attack in the 12$^{th}$ hour of the day could damage the test system for five hours and cause about 3620 kW load shedding in the system. Therefore, it can be concluded that cyber security in electrical grids is a vital issue that must be taken into consideration. In order to improve security in the AMIs, a CADM based on BLSTM and SPRT was proposed to detect FDIA to detect FDIAs on smart meters. In the first phase of the proposed CADM (i.e. prediction phase), the results showed the superiority of the BLSTM over other well-known prediction methods (LSTM, and ANN) such that the BLSM could reach 2.004% MAPE where LSTM and ANN had 2.1928% and 4.1529 respectively. Also, the performance of the CADM tested on a practical data set showed that the proposed model can detect FDIA with only two input samples.